\documentclass[journal]{IEEEtran}
\usepackage[margin=1in]{geometry}

\usepackage[super]{natbib}
\usepackage[pdftex]{graphicx}
\graphicspath{{./figures}}
\DeclareGraphicsExtensions{.pdf,.jpeg,.png,.eps}
\usepackage{amsmath}
\usepackage{array, multirow}
\usepackage{url}
\usepackage{booktabs} 
\usepackage{verbatim} 
\usepackage[final]{pdfpages}

\makeatletter
\renewcommand*{\@biblabel}[1]{\hfill#1.}
\makeatother

\begin{document}

\begin{titlepage}
\vspace*{\fill}
\Large 
\textbf{Diagnostic Accuracy of Content Based Dermatoscopic Image Retrieval with Deep Classification Features}\\
\normalsize 
\textbf{\textit{Tschandl P, Argenziano G, Razmara M, Yap J}}\\
\\
Manuscript submitted for review to the British Journal of Dermatology\\
Final version available at \url{https://doi.org/10.1111/bjd.17189}\\
\\
\\
\large
Citation:\\
\normalsize
\\
@Article\{\\
   tschandl\_cbir2018,\\
   Author="Tschandl, P.  and Argenziano, G.  and Razmara, M.  and Yap, J. ",\\
   Title="{{D}iagnostic {A}ccuracy of {C}ontent {B}ased {D}ermatoscopic {I}mage {R}etrieval with {D}eep {C}lassification {F}eatures}",\\
   Journal="Br J Dermatol",\\
   Year="2018"\\
\}

\vspace*{\fill}
\end{titlepage}

\title{Diagnostic Accuracy of Content Based Dermatoscopic Image Retrieval with Deep Classification Features}

\author{
P~Tschandl\textsuperscript{1,2}, G~Argenziano\textsuperscript{3}, M~Razmara\textsuperscript{4} and J~Yap\textsuperscript{4}\protect\\
}

\maketitle

\textbf{Running head:} Diagnostic accuracy of dermatoscopic image retrieval\\

\textbf{Affiliations:}\\ 
1. School of Computing Science, Simon Fraser University, Burnaby, Canada\\
2. Department of Dermatology, Medical University of Vienna, Vienna, Austria\\
3. Department of Dermatology, University of Campania, Naples, Italy \\
4. MetaOptima Technology Inc., Vancouver, Canada\\

\textbf{Corresponding author:} PD Philipp Tschandl MD PhD, Department of Dermatology,
Medical University of Vienna,
W\"ahringer G\"urtel 18-20,
1090 Vienna,
Austria,
e-mail: first.last$-at-$meduniwien.ac.at\\

\textbf{Funding Sources:} MetaOptima Technology Inc.\\ 

\textbf{Conflict of Interest:} MetaOptima Technology Inc., where Majid Razmara holds the position of Chief Technology Officer, provided access to deep learning hardware, employs J Yap and provides an unrestricted research grant to P Tschandl for conducting a one-year Post-Doc Fellowship at Simon Fraser University. G Argenziano serves as a medical advisor to MetaOptima Technology Inc. MetaOptima Technology offers a CBIR based educational tool to clinicians called \emph{Visual Search} which was not part of presented experiments.\\ 

\textbf{Keywords:} Dermatoscopy, Skin Cancer, Visual Search, Neural Networks, CBIR\\

\noindent\textbf{What's already known about this topic?}
\begin{itemize}
    \item Convolutional Neural Networks (CNN) can detect skin cancer on digital images comparable to dermatologists in experimental settings.
    \item CNNs may be difficult to implement in practice as they commonly output numerical disease probabilities only.
    \item Numerical outputs of intermediate stages of a CNN, referred to as \emph{deep features}, correspond to visual properties on an image.
\end{itemize}
\textbf{What does this study add?}
\begin{itemize}
    \item Content based image retrieval (CBIR) based on deep features can find  visually similar dermatoscopic images.
    \item Retrieving only 16 similar images can achieve the same accuracy as a CNN classifier.
    \item CBIR can enable a CNN to recognise unknown disease classes in new datasets.
\end{itemize}

\vspace{+0.5cm}

\begin{abstract}
\noindent\textbf{\underline{Background:}} Automated classification of medical images through neural networks can reach high accuracy rates but lack interpretability.\\
\textbf{\underline{Objectives:}} To compare the diagnostic accuracy obtained by using content based image retrieval (CBIR) to retrieve visually similar dermatoscopic images with corresponding disease labels against predictions made by a neural network.\\
\textbf{\underline{Methods:}} A neural network was trained to predict disease classes on dermatoscopic images from three retrospectively collected image datasets containing 888, 2750 and 16691 images respectively. Diagnosis predictions were made based on the most commonly occurring diagnosis in visually similar images, or based on the top-1 class prediction of the softmax output from the network. Outcome measures were area under the ROC curve for predicting a malignant lesion (AUC), multiclass-accuracy and mean average precision (mAP), measured on unseen test images of the corresponding dataset.\\
\textbf{\underline{Results:}} In all three datasets the skin cancer predictions from CBIR (evaluating the 16 most similar images) showed AUC values similar to softmax predictions (0.842, 0.806 and 0.852 versus 0.830, 0.810 and 0.847 respectively; p-value$>$0.99 for all). Similarly, the multiclass-accuracy of CBIR was comparable to softmax predictions. Networks trained for detecting only 3 classes performed better on a dataset with 8 classes when using CBIR as compared to softmax predictions (mAP 0.184 vs. 0.368 and 0.198 vs. 0.403 respectively).\\
\textbf{\underline{Conclusions:}} Presenting visually similar images based on features from a neural network shows comparable accuracy to the softmax probability-based diagnoses of convolutional neural networks. CBIR may be more helpful than a softmax classifier in improving diagnostic accuracy of clinicians in a routine clinical setting.
\end{abstract}

\begin{figure*}[ht]
    \centering
    \includegraphics[width=1\textwidth]{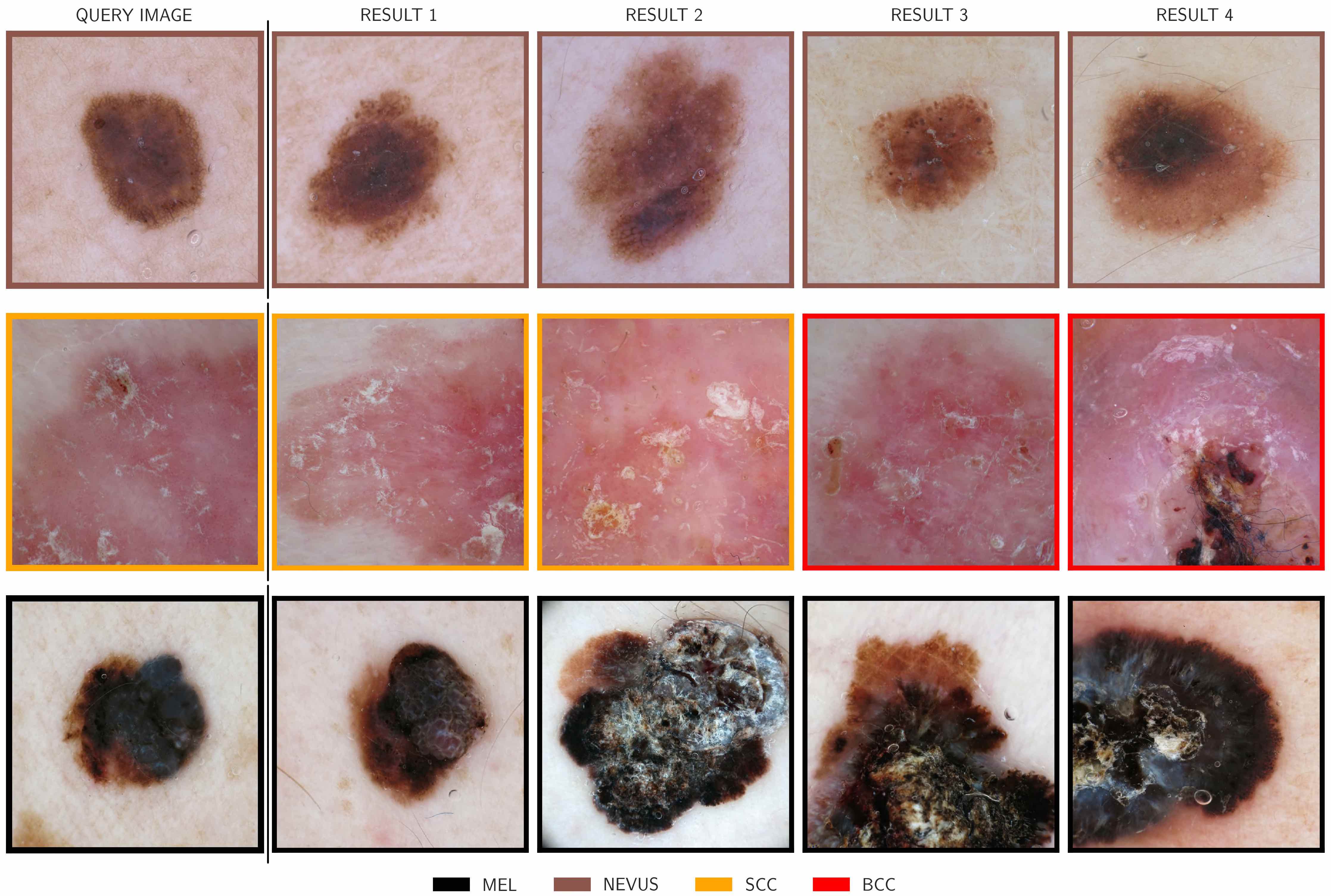}
    \caption{Positive examples of three query images (first column) and corresponding most similar images as found by CBIR. The results show similar dermatoscopic patterns that in the majority correspond to the correct diagnosis.}
    \label{fig:cbirexample}
\end{figure*}

\section{Introduction}

Automated analysis of medical images using neural networks has been used in dermatoscopy for more than a decade \cite{Menzies2005_solarscan, Dreiseitl2007_MelClinPract}, but recently gained attention since groups have reported high accuracy rates with convolutional neural networks (CNN) for skin images \cite{Esteva2017, Han2018} and dermatoscopy \cite{haenssle2018}, as well as for other medical domains such as fundoscopy\cite{Gulshan2016} or chest X-rays \cite{Rajpurkar2017_CheXNet}. CNNs, in brief, are a group of modern and powerful machine learning models that don’t require explicit handcrafted engineering. Rather, they learn to detect visual elements such as colors, shapes and edges by themselves, and combine detections of those internally to a prediction. The only thing needed for them, apart from computing power, is a large number of images and labels to train them, where the labels correspond to the diagnosis in the medical field.\\
Implementing automated classification models, like a CNN, that output probabilities of diagnoses, or the most probable diagnosis, is deemed desirable for a number of reasons within a health care system. Using patient-based methods could ultimately reduce the need for physicians in areas of scarcity and reduce burden on the health care system, but are highly problematic in regard to regulations and safety. A more realistic approach is having decision support systems available to non-specialised physicians that may be easier to implement and have the potential to increase their diagnostic accuracy and decrease referral rates. Integrating classification systems into a specialists’ clinical workflow may increase efficiency and free them from spending a large amount of time on easy to diagnose cases.\\

While these effects are undeniably positive, real-world settings can be problematic for classifiers that output the probability of a diagnosis. Accuracy rates for specified cutoffs are commonly reported in experimental settings on digital images with a verification bias, as mainly pathologically verified diagnoses are  deemed the gold standard for ground-truth labels \cite{Codella2017_ISIC}. Even in sets using expert evaluations as “labels”, the included cases may not inherit all or enough representations of common banal skin diseases \cite{HAM10000}. Specialised centers may not bother photo-documenting such common cases due to the additional time required, and given their obvious diagnosis to an expert.\\ Apart from imperfect accuracy rates of neural networks, unforeseen problems can arise in practical use. This is exemplified by an earlier clinical study using an automated skin lesion classifier where melanomas where missed simply because they were not photographed by the user \cite{Dreiseitl2007_MelClinPract}.\\ 
Lastly, classifications of CNNs can be prone to adversarial examples \cite{Finlayson2018_medadversarials} raising questions of liability in misdiagnoses of such systems, or falsely vindicating skin lesion removal on insurance funds for cosmetic or financial incentives.\\

A solution for these problems is to keep physicians "in the loop" \cite{Girardi2016_HITL} for automated diagnoses. Classification systems could run in the background analyzing images to bring the ones of most concern to a doctor's attention more quickly. These systems could also be used to continuously audit previously diagnosed cases where disagreements between the automated classifier and physician can be flagged and recommended for review. For a successful human-machine collaboration it is key to know why a system makes a specific diagnosis, options being visual question answering or automated captioning \cite{Park2016_XAIVQA}. For all these systems it is left to the discretion of the user to interpret the results and decide whether they are correct.
Herein we explore a different, intuitive and transferrable approach for 'explainable' artificial intelligence (AI), called content based image retrieval (\emph{CBIR}). With CBIR, the user presents an unknown query image to a system, and cases with similar visual features are automatically retrieved and displayed from a database. Example queries and results of automatically retrieved similar images are shown in Figure \ref{fig:cbirexample}.\\

With the increased performance of convolutional neural networks in regard to classification, previous work has found that those networks also learn filters that correspond to visual elements of an image in later layers of a CNN \cite{Piplani2018_DeepSeek}. In other words, one set of filters in a CNN could for example respond to whether a brown network is visible, and another one could respond to a group of blue clods. With many filters present in a CNN, and many ways to combine them as an image moves through the network, it is an active research area to try and understand what set of filters correspond to an exact given visual structure. However, even without knowing what exact filter detects which structure, taken altogether they can be expressed as row of simple numbers (called a “feature vector” or “deep features”), representing all visual elements in an image. By comparing how similar these collected numbers of two images are, one can match faces\cite{Parkhi2015_DeepFace}, or retrieve visually similar medical data such as histopathologic images\cite{Xiaoshuang2018_binaryhash}.
Recently, Kawahara et al.\cite{Kawahara2018_7Point} used such extracted features of a multi-modality network to query a database for similar images and found it had high sensitivity (94\%) but low specificity (36\%) for detecting melanoma (73\% and 79\% respectively for a different diagnostic cutoff).

The goals of this study are:
\begin{itemize}
    \item To evaluate whether CBIR based on deep features of a neural network, trained for classification, can provide a comparable diagnostic accuracy as its softmax probabilities.
    \item To determine how many similar images may be practically needed.
    \item To determine whether a CBIR system is transferrable to different datasets.
\end{itemize}

\section{Methods}

\begin{table*}[!htbp]
\centering
\caption{Presentation of used study datasets with numbers of included diagnoses. EDRA and ISIC2017 contain the same disease classes, whereas the private (PRIV) dataset contains 8 different diagnoses.}
\scalebox{0.85}{
\begin{tabular}{lrrrrrrrrr}
\toprule
Dataset &  Total & angioma & bcc & bkl & df &  inflammatory & mel & nevus & scc \\
\midrule
EDRA    &   888 (100\%) &      - &     - &    69 (7.8\%) &    - &   - &   247 (27.8\%) &   572 (64.4\%) &       - \\
ISIC2017&  2750 (100\%) &      - &     - &   386 (14.0\%) &    - &   - &   521 (18.9\%) &  1843 (67.0\%) &       - \\
PRIV    & 16691 (100\%) &    203 (1.2\%) &  3842 (23.0\%) &  1368 (8.2\%) &  206 (1.2\%) & 566 (3.4\%) &  2276 (13.6\%) &  5941 (35.6\%) &  2289 (13.7\%) \\
\bottomrule
\end{tabular}
}
\label{tbl:datasets}
\end{table*}

\subsection{Datasets}
We compare diagnostic performance of a CBIR system to neural networks using the following 3 datasets:
\begin{itemize}
\item EDRA: A large collection of dermatoscopic images was published alongside the \emph{Interactive Atlas of Dermoscopy}\cite{Argenziano2000_EDRA}. We filtered the dataset to contain only diagnoses with more than 50 examples and are consistent with the ISIC2017 dataset. 20\% of the images, randomised and stratified by diagnosis of cases, were split as a test-set to evaluate our method on. Of the remaining cases, 20\% were used as validation during training to fit network training parameters.
\item ISIC2017: The International Skin Imaging Collaboration (ISIC) 2017 challenge for melanoma recognition published a convenience dataset of dermatoscopic images with fixed training, validation and test splits \cite{Codella2017_ISIC}. The diagnoses included in the dataset are melanoma (\emph{mel}), nevi (\emph{nevus}) and seborrheic keratoses (\emph{bkl}). 
\item PRIV: We gathered dermatoscopic images that were consecutively collected at a single skin cancer center between 2001 and 2016 for clinical documentation including pathologic and clinical diagnoses \footnote{Ethics review board waiver from Ospedaliera di Reggio Emilia, Protocol No. 2011/0027989}. We excluded diagnoses with less than 150 examples, which resulted in inclusion of the following diagnoses: angioma (incl. angiokeratoma), \emph{bcc} (basal cell carcinoma), \emph{bkl} (seborrheic keratoses, solar lentigines and lichen planus-like keratoses), \emph{df} (dermatofibromas),  \emph{inflammatory} lesions (including dermatitis, lichen sclerosus, porokeratosis, rosacea, psoriasis, lupus erythematosus, bullous pemphigoid, lichen planus, granulomatous processes and artifacts), \emph{mel} (all types of melanomas), \emph{nevus} (all types of melanocytic nevi) and \emph{scc} (squamous cell carcinomas, actinic keratoses and bowen's disease). We performed splitting in the same manner as for the \emph{EDRA} dataset for cases with a pathologic diagnosis. Cases that had no pathologic diagnosis but an expert rating were included only in the training set.
\end{itemize}
For all datasets, the training-set also represents the pool for images possibly retrieved by the tested CBIR systems. We avoided same-lesion images spread between training-, validation- and test-set. Complete dataset numbers are shown in Table \ref{tbl:datasets}.\\

\subsection{Network architecture and training}
In all experiments we use a ResNet-50 architecture\cite{He2015} with network parameters obtained through training on the ImageNet\cite{ILSVRC2015} dataset, which contains $>$1 million images of 1000 different objects of daily life. This pretraining enables the ResNet-50 architecture to recognize general shape, edge and colour combinations, and reduces the training time needed to adapt it to our specialized task of dermatoscopic image classification. 
Depending on the dataset used for a given experiment we modify the size of the last \emph{fully-connected layer} in the CNN to match the number of classes present respectively, and fine-tune the network. This \emph{fully connected layer} provides the probabilty output for every diagnosis, and because this layer processes its numerical input with the \emph{softmax} function, we refer to its output as 'softmax prediction'. As compared to Han et al.\cite{Han2018} we don't define diagnosis-specific thresholds, but rather take the diagnosis with the highest probability value as the final diagnosis prediction. Further training implementation details are given in the Supplementary File.

\subsubsection{CBIR}
For all images in the retrieval image set we pass them through the CNN, and collect the output of the deepest feature layer ("pool5") as our feature vector. This vector consists of 2048 numbers that represent visual features of an image. By calculating the cosine similarity of two such vectors, we get a single number ranging between 0 and 1 corresponding to how 'similar' features in two images are. In other words, the cosine similarity of two images describes in a single number how similar the visual elements of two images are. So, to obtain the most visually similar images to a query in this study, we calculate its cosine similarity to every other image in a dataset and sort them by the resulting value. In order to be able to compare CBIR with softmax predictions, we collect the k most similar lesions for every query and regard the frequency of their corresponding disease labels as their probability. For example, if 4 of 5 similar images are a melanoma and one is a nevus, we regard melanoma probability as 0.8 and nevus probability as 0.2.

\subsection{Metrics and Statistics}
The following metrics are calculated for evaluating diagnostic accuracy, where all retrieved images had the same weight during retrieval except for solving ties of specific diagnoses:
\begin{itemize}
    \item Area under the ROC curve for detecting skin cancer ($AUC$): Percent of malignant retrieval cases (CBIR) or the sum of probabilities of malignant classes (Softmax) are used to calculate ROC curves. Sensitivity and specificity values are likewise calculated for detecting skin cancer with fixed cutoffs of needed malignant examples / probabilities returned (25\% ($Sens@0.25$ and $Spec@0.25$) and 50\% ($Sens@0.5$ and $Spec@0.5$) of retrievals). Due to the lack of other malignant classes, this value is equal to the AUC to detect melanoma when testing on EDRA and ISIC2017 datasets.
    
    \item Multi-class Accuracy ($Accuracy$): Percentage of all correct specific predictions, where the prediction is made for the class with the highest probability (Softmax) or most commonly retrieved (CBIR) examples. To avoid tied predictions with CBIR, a minimal linear weighting based on retrieval order (1.00-0.99 distributed evenly along $k$ retrieved images) is applied during counting.
    
    \item Multi-class Mean Average Precision ($mAP$): Briefly, average precision scores for every test-set class are macro-averaged as implemented by \cite{scikit-learn}, where prediction scores were obtained by either the frequency of the query class in CBIR retrievals or softmax prediction scores. A more detailed description is given in Supplementary File 1.
    
\end{itemize}

\begin{figure*}[ht]
    \centering
    \includegraphics[width=1\textwidth]{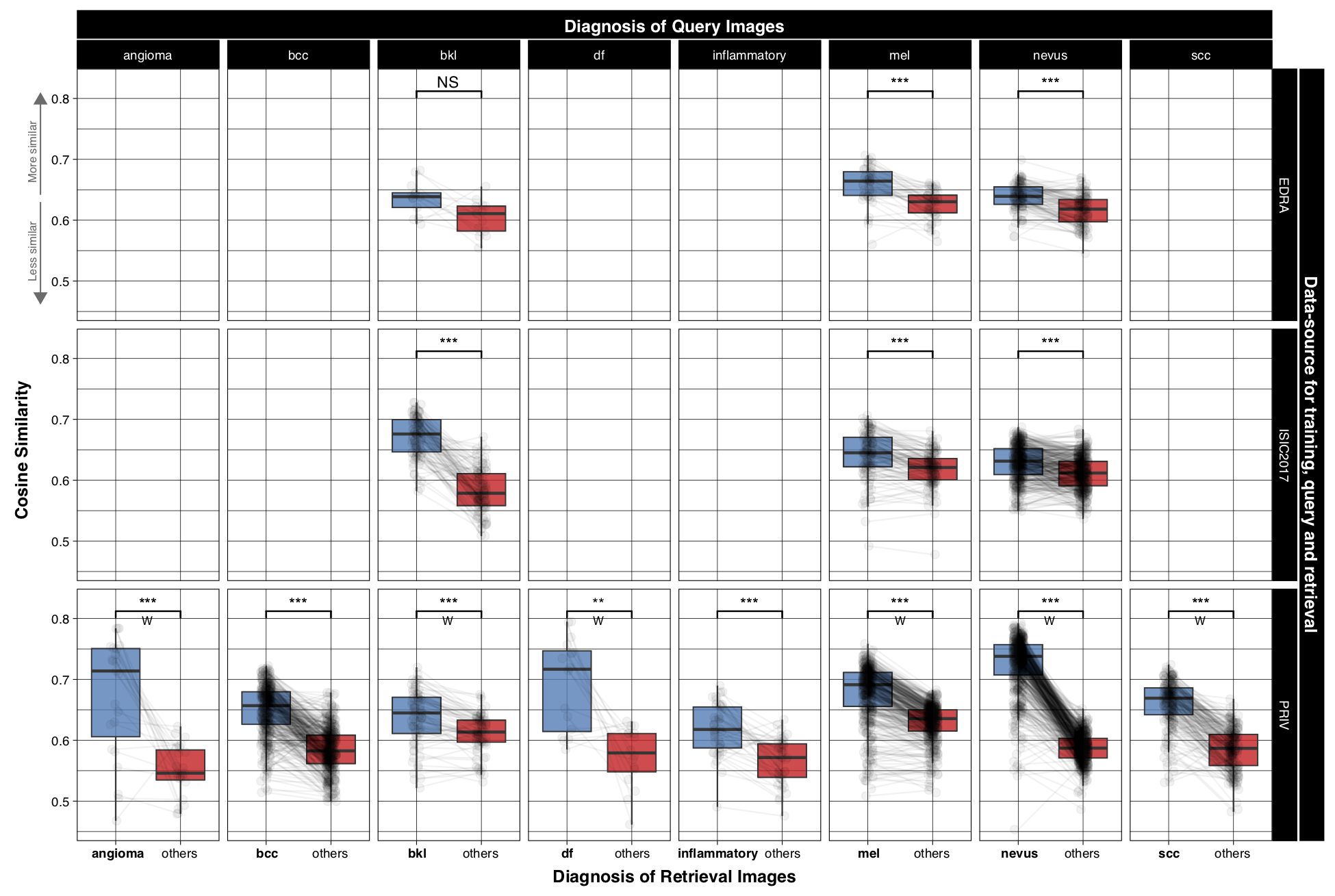}
    \caption{Measured visual similarity (cosine similarity) of images with the same diagnoses (blue) compared to others (red) in a dataset. Images of the same diagnoses are significiantly rated higher in almost any subgroup, showing automated measurements of visual similarity can differentiate between diagnoses within a retrieval dataset. Lines are drawn between values for a single query image, and rows denote dataset used for training, queries and image retrieval. Comparing differences was performed with a paired t-test or a paired Wilcoxon signed rank test (\emph{W}). NS p-value$>$0.05, * p-value$<$0.05, ** p-value$<$0.01, *** p-value$<$0.001.}
    \label{fig:similarity_percase}
\end{figure*}

Experiments as well as raw data computation and visualisation are performed with python (PyTorch\cite{pytorch}, sklearn\cite{scikit-learn} and matplotlib\cite{matplotlib}) and R Statistics \cite{R_Statistics, ggplot2}. As testing all combinations of CBIR cutoffs (restricted to up to 32 images), datasets and metrics would result in too many comparisons, we restricted formal statistical tests comparing diagnostic metrics to the AUC of ROC detecting skin cancer when retrieving 2, 4, 8, 16, and 32 images which we believe is a clinically meaningful evaluation. ROC curves are computed using pROC\cite{Xavier2011_pROC} and compared using the DeLong method \cite{Delong1988}. Paired t-tests are used to compare cosine similarity values after checking for approximate normality. In case of a violation, paired Wilcoxon signed rank test is used instead. A two-sided p-value of $<$0.05 is regarded statistically significant. 95\%CI values of ROC curves as well as sensitivity and specificity at specified cutoffs are calculated with 2000 bootstrapped replicates. All p-values are reported adjusted for multiple testing with the Holm method \cite{Holm1979} unless otherwise specified. Correction for multiple testing was stopped after the first non-rejection of the null-hypothesis, and therefore no adjusted p-values reported for the remaining comparisons.


\section{Results}

\subsection{Same-source CBIR and classification}

The mean cosine similarities of all retrieval images for all queries of the same data-source were 0.631 (95\%CI: 0.628-0.634; EDRA), 0.623 (95\%CI: 0.621-0.625; ISIC2017) and 0.638 (95\%CI: 0.635-0.640; PRIV). Retrieval images with the same diagnosis had a significantly higher similarity value to a query image compared to those of different classes (0.667 (95\%-CI: 0.665-0.669) vs. 0.601 (95\%-CI: 0.600-0.603); p$<$0.001). Subgroup analyses likewise revealed significant differences for every diagnosis within every dataset (see Figure \ref{fig:similarity_percase}). For accuracy calculations below, the $k$ most similar retrieval images were collected for every query, and the most frequent occurring disease label counted as the prediction. 

\begin{figure*}[h]
    \centering
    \includegraphics[width=1\textwidth]{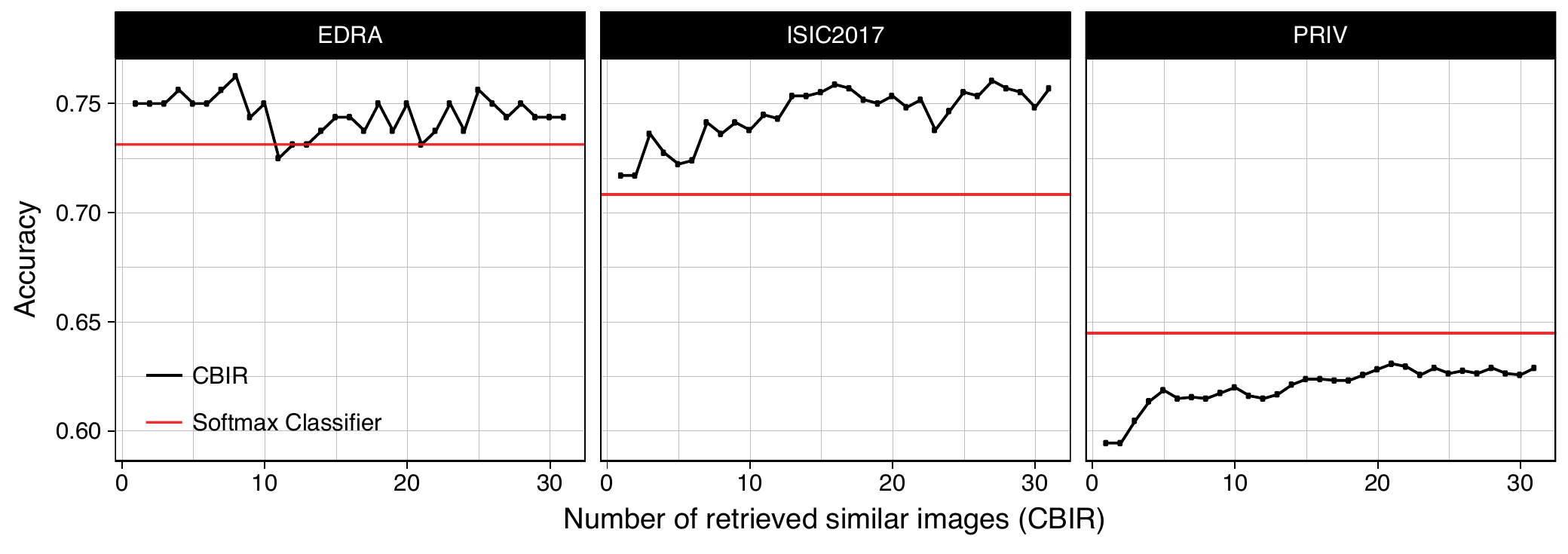}
    \caption{Frequency of correct specific diagnoses (\emph{Accuracy}) made within each dataset by either softmax based predictions (red), or CBIR with a different number of retrieved similar images (black). Retrieval of already few images is performing better in the 3-class datasets (EDRA, ISIC2017), whereas in the 8-class (PRIV) dataset it takes over 20 images to approximate softmax based accuracy.}
    \label{fig:sameset}
\end{figure*}

Using these ranked images for diagnostic predictions was able to approximate a classic softmax-based classifier with only few retrieval cases in regard to multi-class accuracy (Figure \ref{fig:sameset} and Table \ref{tbl:auccancer}). For the two datasets containing only 3 classes, CBIR outperformed the softmax-based classification and had the highest accuracy when retrieving 8 (EDRA, accuracy$=$0.762) and 16 similar cases (ISIC2017, accuracy$=$0.759), whereas in the PRIV dataset the best result with 32 retrievals (accuracy 0.629) was still below the corresponding softmax accuracy of 0.645. As can be seen in Figure \ref{fig:sameset}, using more than 16 retrieved images did not consistently improve accuracy of CBIR. In all three datasets, showing only two retrieved images resulted in decreased performance in detecting skin cancer as measured by the AUC, where the difference was significant for the 8-class dataset (EDRA 0.782 vs. 0.830, p$=$1.0; ISIC2017 0.760 vs. 0.810, p$=$0.073; PRIV 0.791 vs 0.847, p$<$0.001).

\begin{table*}[htbp]
\centering
\caption{Intra-dataset performance metrics. AUC is calculated for detecting any malignant skin tumor in the corresponding dataset. Sensitivity (\emph{Sens}) and Specificity (\emph{Spec}) are calculated at 25\% and 50\% retrieved malignant cases (CBIR) or predicted probability of malignancy (Softmax). p-values, provided original and as \emph{p-value (adj.)} with correction for multiple testing by the method of Holm\cite{Holm1979}, denote difference of CBIR based AUC values to Softmax based ones. $\ddagger$ signifies non-evaluated comparisons after correction for multiple testing. Numbers in brackets represent 95\% confidence intervals.}
\scalebox{0.8}{
\begin{tabular}{llrlllllrr}
\toprule
\textbf{Dataset} & \textbf{CBIR (k)} & $Accuracy$ &     $Sens@0.25$ &                 $Spec@0.25$ &                  $Sens@0.5$ &                  $Spec@0.5$ &                       $AUC$ & $p-value$ & $p-value (adj.)$ \\
\midrule
\multirow{6}{*}{\textbf{EDRA}} & \textbf{2} &  0.750 &   72.7 (59.1-84.1) &  \textbf{78.4 (70.7-85.3)} &  \textbf{72.7 (59.1-86.4)} &  78.4 (70.7-85.3) &  0.782 (0.703-0.861) & 0.151 &  $\ddagger$ \\
     & \textbf{4} &                              0.756 &  \textbf{88.6 (79.5-97.7)} &    64.7 (56-73.3) &    63.6 (50.0-77.3) &  86.2 (79.3-92.2) &    0.830 (0.760-0.900) & $>$0.99 &  $\ddagger$ \\
     & \textbf{8} &                            \textbf{0.762} &   86.4 (75.0-95.5) &  70.7 (62.1-79.3) &  56.8 (40.9-70.5) &  89.7 (84.5-94.8) &  \textbf{0.850 (0.784-0.916)} & 0.342 &  $\ddagger$ \\
     & \textbf{16} &                        0.744 &  84.1 (72.7-93.2) &  68.1 (59.5-76.7) &  52.3 (38.6-65.9) &  89.7 (83.6-94.8) &  0.842 (0.776-0.908) & 0.491 &   $\ddagger$ \\
     & \textbf{32} &                           0.744 &    86.4 (75.0-95.5) &  69.8 (61.2-77.6) &  47.7 (31.8-61.4) &  \textbf{92.2 (87.1-96.6)} &  0.844 (0.776-0.912) & 0.499 &   $\ddagger$ \\ \cline{2-10}
     & \textbf{Softmax} &                      0.731 &   77.3 (65.9-88.6) &  75.9 (68.1-83.6) &  61.4 (47.7-75.0) &  84.5 (77.6-91.4) &  0.830 (0.759-0.901) & - &     - \\
\cline{1-10}
\multirow{6}{*}{\textbf{ISIC2017}} & \textbf{2} &   0.717 &   66.7 (58.1-75.2) &    83.2 (80-86.7) &  \textbf{66.7 (58.1-75.2)} &  83.2 (79.7-86.7) &  0.760 (0.713-0.807) & 0.006 &  0.073 \\
     & \textbf{4} &                                  0.727 &   \textbf{76.1 (68.4-83.8)} &  71.9 (67.5-76.0) &  53.8 (44.4-63.2) &  89.5 (86.5-92.4) &  0.785 (0.737-0.833) & 0.118 &  $\ddagger$ \\
     & \textbf{8} &                                 0.736 &   70.1 (61.5-78.6) &  \textbf{78.4 (74.7-82.1)} &  50.4 (41.9-59.8) &  90.4 (87.8-93.0) &  0.798 (0.751-0.845) & 0.431 &  $\ddagger$ \\
     & \textbf{16} &                                  \textbf{0.759} &   70.9 (62.4-78.6) &  77.6 (73.9-81.3) &    50.4 (41-59.8) &  92.8 (90.4-95.2) &  \textbf{0.806 (0.759-0.853)} & 0.785 &  $\ddagger$ \\
     & \textbf{32} &                                   0.753 &    68.4 (59.8-76.9) &  77.3 (73.6-81.0) &  41.0 (32.5-49.6) &  \textbf{94.3 (92.2-96.3)} &  0.799 (0.751-0.846) & 0.354 &  $\ddagger$ \\ \cline{2-10}
     & \textbf{Softmax} &                             0.708 &   70.9 (62.4-79.5) &  74.9 (70.8-78.6) &  60.7 (52.1-69.2) &  86.7 (83.4-89.5) &  0.810 (0.765-0.854) & - &    - \\
\cline{1-10}
\multirow{6}{*}{\textbf{PRIV}} & \textbf{2} &       0.594 &  82.7 (80.2-85.1) &  66.8 (63.1-70.6) &  \textbf{82.7 (80.2-85.3)} &  67.0 (63.1-70.6) &    0.791 (0.770-0.813) & $<$0.001 &  $<$0.001 \\
     & \textbf{4} &                              0.614 &   \textbf{89.3 (87.2-91.3)} &  54.4 (50.5-58.3) &  79.4 (76.8-81.9) &  73.9 (70.3-77.3) &  0.822 (0.802-0.843) & 0.002 &  0.032 \\
     & \textbf{8} &                              0.615 &   87.9 (85.9-89.9) &  60.6 (56.9-64.4) &    74.7 (72-77.4) &  77.9 (74.8-81.2) &  0.843 (0.823-0.862) & 0.597 &  $\ddagger$ \\
     & \textbf{16} &                              0.624 &   87.4 (85.2-89.5) &    63.9 (60-67.8) &  74.2 (71.2-76.9) &  81.2 (78.1-84.3) &  0.852 (0.833-0.871) & 0.456 &  $\ddagger$ \\
     & \textbf{32} &                               \textbf{0.629} &   87.2 (84.9-89.3) &  \textbf{66.2 (62.4-69.8)} &  73.2 (70.5-75.9) &  \textbf{82.2 (78.9-85.1)} &    \textbf{0.859 (0.840-0.878)} & 0.072 &  $\ddagger$ \\ \cline{2-10}
     & \textbf{Softmax} &                           0.645 &   87.7 (85.5-89.7) &  62.6 (58.7-66.3) &  75.3 (72.5-78.0) &  79.7 (76.6-82.7) &  0.847 (0.827-0.867) & - &    - \\
\bottomrule
\end{tabular}
}
\label{tbl:auccancer}
\end{table*}

\begin{figure}[h]
    \centering
    \includegraphics[width=0.52\textwidth]{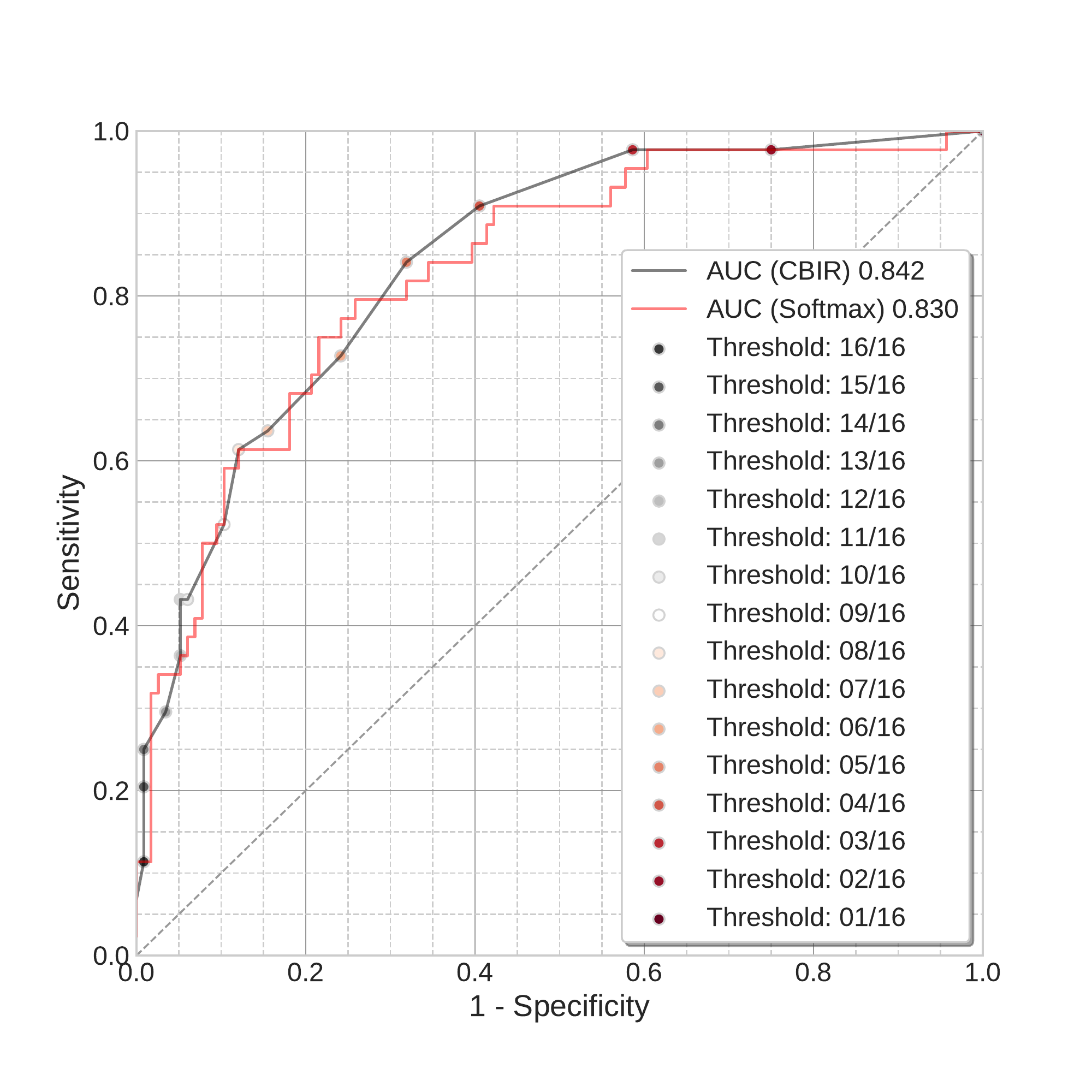}
    \caption{ROC for detecting melanoma when retrieving 16 similar images with CBIR (grey), showing different thresholds of needed malignant retrieval images ("predict melanoma when \emph{x} of 16 retrieved images are melanomas"), as well as with softmax based probabilities (red). Network training-, query- and retrieval- images are from EDRA.}
    \label{fig:auccancer}
\end{figure}

Figure \ref{fig:auccancer} shows the ROC curve of the EDRA intra-dataset evaluation when fixing the CBIR output to 16 images, where disregarding a small frequency of malignant cases in the images doesn't change sensitivity substantially. Fixing the outputs to 16 cases and labeling a query case "malignant" if at least 25\% of retrievals show a malignant lesion, results in a sensitivity of 84.1\% at a specificity of 68.1\% in the EDRA dataset, 70.9\% and 77.6\% in the ISIC2017, and 87.4\% and 63.9\% in the PRIV dataset respectively (Table \ref{tbl:auccancer}).

\subsection{New-source classification}
\begin{figure*}[ht]
    \centering
    \includegraphics[width=1\textwidth]{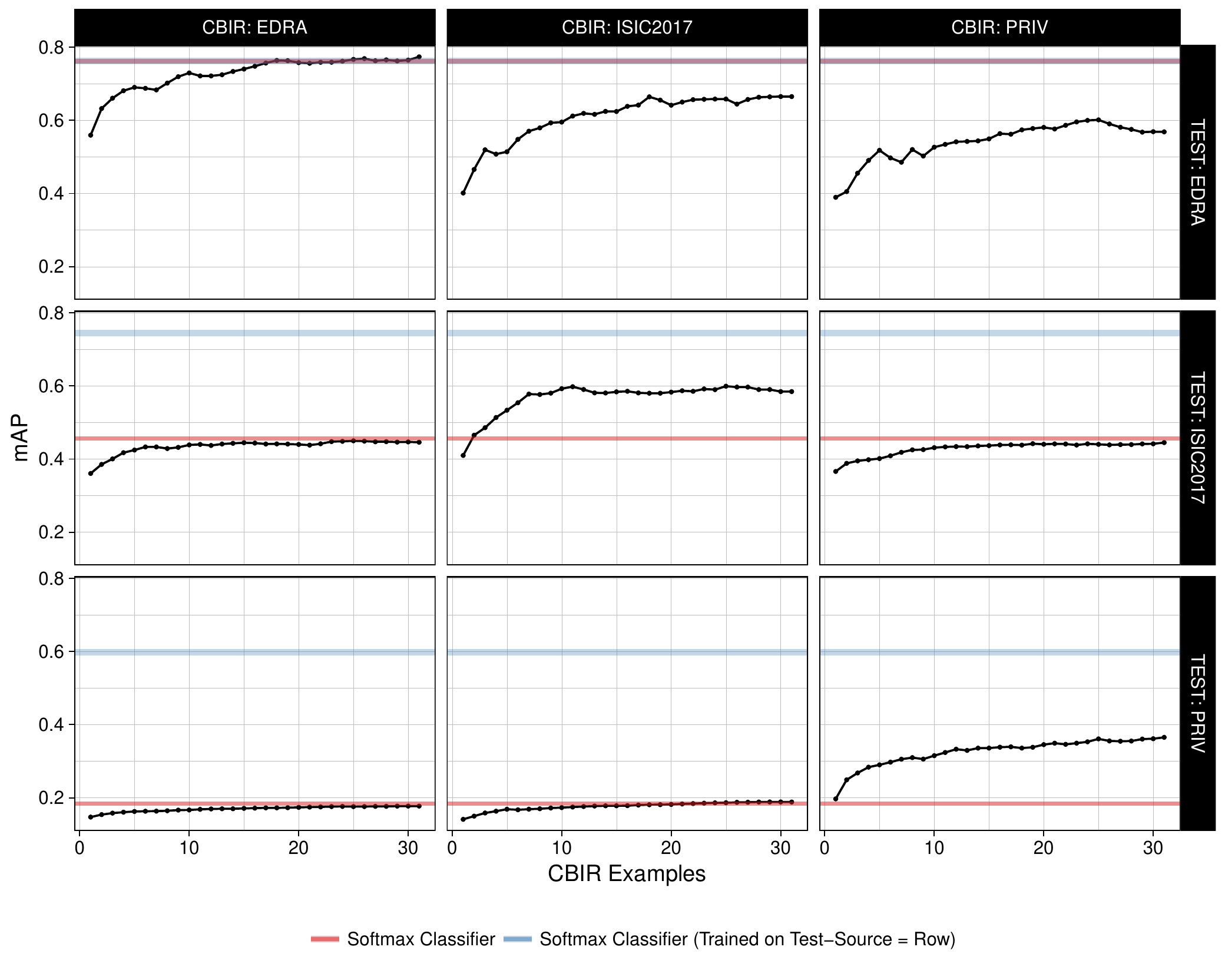}
    \caption{Mean Average Precision (\emph{mAP}) of a ResNet-50 network trained on EDRA dataset images. Predictions are made either through  softmax probabilities (red line) or class-frequencies of CBIR (black). Softmax predictions perform bad on predicting PRIV dataset images, as the network as not able to predict 5 of the 8 classes in any case (first to columns, bottom row). CBIR retrieving from EDRA and ISIC2017 suffers from the same shortcomings, but is able to predict better when using PRIV-source retrieval images (bottom right). In general, CBIR performs best when using retrieval images from the same source as the test images (descending diagonal), and here performs better on new data than softmax predictions. Re-training the network on those new source images (blue) in turn outperforms CBIR again.}
    \label{fig:diffset}
\end{figure*}

\begin{figure*}[ht]
    \centering
    \includegraphics[width=1\textwidth]{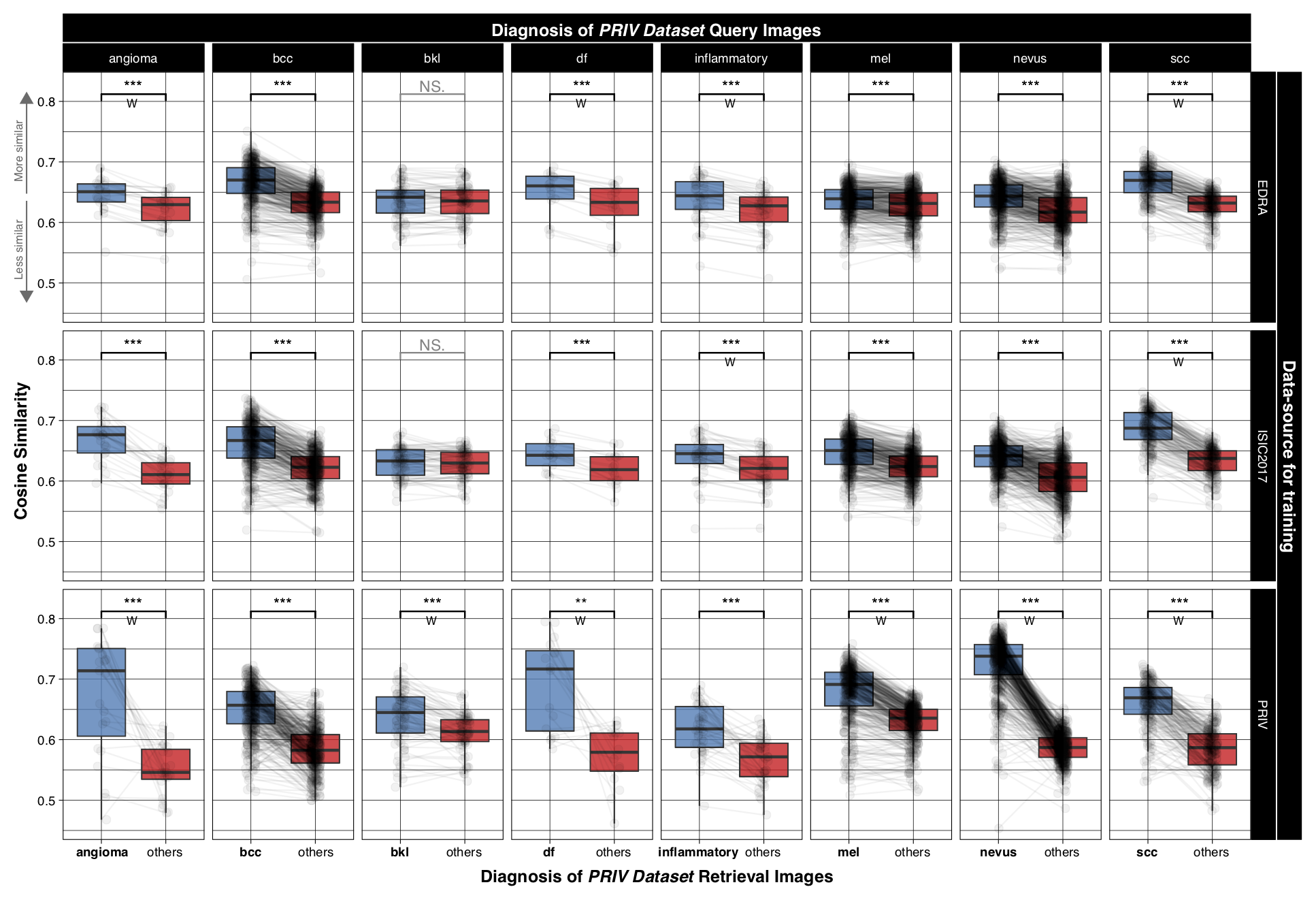}
    \caption{Mean cosine similarities of PRIV retrieval images with the same (blue) or different (red) diagnosis for the correponding PRIV query images. Cosine similarity is calculated by feature extraction of ResNet-50 networks trained for classification on different training datasets (rows). Compared to the PRIV-trained network, those trained on different sources (row EDRA and ISIC2017) transfer their ability to distinguish specific diagnoses through visual similary except for bkl cases. Lines are drawn between values for the same query image. \emph{W} denotes use of paired Wilcoxon signed rank test instead of paired t-test. NS p-value$>$0.05, * p-value$<$0.05, ** p-value$<$0.01, *** p-value$<$0.001, grey indicators denote non-adjusted p-values as these comparisons were omitted during correction for multiple testing (see statistics section).}
    \label{fig:sup_diffsetsimilarity}
\end{figure*}

Figure \ref{fig:diffset} and Table \ref{tbl:mapdiff} show mean average precision values of networks trained and tested on different datasets, with different CBIR resource databases used. In other words, the images to be diagnosed, the images a CNN retrieves similar cases from, and the images the CNN was trained on can all originate from different sources. Softmax based predictions from 3-class-trained networks (EDRA \& ISIC2017) perform worse on predicting the 8-class dataset (PRIV) with $mAP$ values of 0.184 and 0.198 respectively. Using the target source as a CBIR resource improved $mAP$ to up to 0.368 and 0.403 respectively. This is because previously 'unknown' classes can still be retrieved as those networks transfer the ability to distinguish diagnoses through visual similarity (see Figure \ref{fig:sup_diffsetsimilarity}). The best CBIR performance is obtained with combinations where training, testing and resource are from the same source.

\begin{table}[htbp]
\centering
\caption{Mean average precision between datasets. TRAIN denotes dataset the ResNet-50 architecture was trained on, TEST the origin of test images, and CBIR origin of retrieval images. While CBIR is able to approximate Softmax-based predictions between the 3-class datasets (EDRA and ISIC2017) when using same-source TEST and CBIR sets, it outperforms 3-class trained networks on the 8-class PRIV dataset as it is able to recognise unseen classes through the larger resource dataset.}
\scalebox{0.55}{
\begin{tabular}{lllrrrrrr}
\toprule
\textbf{TRAIN} & \textbf{TEST} & \textbf{CBIR}  & $CBIR_2$ & $CBIR_4$ & $CBIR_8$ & $CBIR_{16}$ & $CBIR_{32}$ &  $Softmax$ \\
\midrule
\multirow{9}{*}{\textbf{EDRA}} & \multirow{3}{*}{\textbf{EDRA}} & \textbf{EDRA} &  0.632 &  0.681 &  0.702 &  0.748 &  0.775 &  \multirow{3}{*}{\textbf{0.761}} \\
     &      & \textbf{ISIC2017} &  0.466 &  0.507 &  0.579 &  0.638 &  0.662 &     \\
     &      & \textbf{PRIV} &  0.405 &  0.490 &  0.520 &  0.563 &  0.573 &  \\
\cline{2-9}
     & \multirow{3}{*}{\textbf{ISIC2017}} & \textbf{EDRA} &  0.385 &  0.417 &  0.429 &  0.444 &  0.444 & \multirow{3}{*}{\textbf{0.456}} \\
     &      & \textbf{ISIC2017} &  0.465 &  0.513 &  0.576 &  0.585 &  0.582 &    \\
     &      & \textbf{PRIV} &  0.388 &  0.398 &  0.425 &  0.438 &  0.445 &    \\
\cline{2-9}
     & \multirow{3}{*}{\textbf{PRIV}} & \textbf{EDRA} &  0.154 &  0.161 &  0.165 &  0.172 &  0.177 &  \multirow{3}{*}{\textbf{0.184}} \\
     &      & \textbf{ISIC2017} &  0.150 &  0.163 &  0.170 &  0.179 &  0.188 &   \\
     &      & \textbf{PRIV} &  0.249 &  0.284 &  0.310 &  0.338 &  0.368 &    \\
\cline{1-9}
\cline{2-9}
\multirow{9}{*}{\textbf{ISIC2017}} & \multirow{3}{*}{\textbf{EDRA}} & \textbf{EDRA} &  0.524 &  0.591 &  0.583 &  0.624 &  0.604 & \multirow{3}{*}{\textbf{0.524}} \\
     &      & \textbf{ISIC2017} &  0.410 &  0.448 &  0.487 &  0.488 &  0.512 &    \\
     &      & \textbf{PRIV} &  0.374 &  0.416 &  0.441 &  0.453 &  0.459 &    \\
\cline{2-9}
     & \multirow{3}{*}{\textbf{ISIC2017}} & \textbf{EDRA} &  0.376 &  0.403 &  0.459 &  0.504 &  0.537 & \multirow{3}{*}{\textbf{0.745}} \\
     &      & \textbf{ISIC2017} &  0.583 &  0.654 &  0.697 &  0.725 &  0.734 &   \\
     &      & \textbf{PRIV} &  0.405 &  0.423 &  0.439 &  0.468 &  0.483 &    \\
\cline{2-9}
     & \multirow{3}{*}{\textbf{PRIV}} & \textbf{EDRA} &  0.149 &  0.158 &  0.167 &  0.175 &  0.182 & \multirow{3}{*}{\textbf{0.198}} \\
     &      & \textbf{ISIC2017} &  0.159 &  0.172 &  0.183 &  0.191 &  0.200 &   \\
     &      & \textbf{PRIV} &  0.269 &  0.316 &  0.377 &  0.389 &  0.403 &   \\
\cline{1-9}
\cline{2-9}
\multirow{9}{*}{\textbf{PRIV}} & \multirow{3}{*}{\textbf{EDRA}} & \textbf{EDRA} &  0.514 &  0.597 &  0.637 &  0.647 &  0.640 &  \multirow{3}{*}{\textbf{0.641}} \\
     &      & \textbf{ISIC2017} &  0.434 &  0.465 &  0.498 &  0.540 &  0.566 &   \\
     &      & \textbf{PRIV} &  0.543 &  0.552 &  0.582 &  0.597 &  0.629 &    \\
\cline{2-9}
     & \multirow{3}{*}{\textbf{ISIC2017}} & \textbf{EDRA} &  0.371 &  0.403 &  0.434 &  0.458 &  0.475 &  \multirow{3}{*}{\textbf{0.551}} \\
     &      & \textbf{ISIC2017} &  0.543 &  0.596 &  0.649 &  0.667 &  0.688 &   \\
     &      & \textbf{PRIV} &  0.419 &  0.446 &  0.468 &  0.498 &  0.528 &   \\
\cline{2-9}
     & \multirow{3}{*}{\textbf{PRIV}} & \textbf{EDRA} &  0.152 &  0.161 &  0.167 &  0.171 &  0.177 &  \multirow{3}{*}{\textbf{0.598}} \\
     &      & \textbf{ISIC2017} &  0.158 &  0.169 &  0.181 &  0.188 &  0.197 &   \\
     &      & \textbf{PRIV} &  0.405 &  0.472 &  0.517 &  0.545 &  0.568 &   \\
\bottomrule
\end{tabular}
}
\label{tbl:mapdiff}
\end{table}

\section{Discussion}
Current convolutional neural network (CNN) classifiers perform well but commonly behave as black-boxes during inference and preclude meaningful integration of their findings to a clinical decision process. Having an intuitive, 'explainable', output of an automated classifier which complements - rather than overrides - a clinical decision process may be more desirable and can enhance efficient use of health care workers. Compared to other techniques for explainable AI \cite{Holzinger2017_explainable} such as image captioning and visual question answering \cite{Park2016_XAIVQA}, we hypothesize that showing similar cases with their ground truth may be even more intuitive. Similar images found by CBIR further comprehensibly reveal the knowledge base of a network decision and may conceive when not to trust the automated system. More specifically, if users notice retrieved cases look nothing like the query image, they could intuitively decide the CNN cannot help in that case. 
Herein we show that CBIR can perform on par with softmax-based predictions of a ResNet-50 network on accuracy of skin cancer detection, as well as multi-class accuracy and mean average precision (Table \ref{tbl:auccancer}).\\
We describe reasonably good metrics for formal evaluation of a CBIR system, but more current architectures may be able to reach even higher accuracy. We hypothesise, that with increasing accuracy of a network, accuracy of CBIR will rise accordingly. The true advantage of CBIR may lie in that a human reader can pick the most fitting and relevant examples out from the provided image-subset and is not restricted to the strict counting and weighting used for calculations in this manuscript. We suspect having such a 'human-in-the-loop' would give a much higher diagnostic precision in practice, which should be subject to future studies.

Deep learning literature dealing with image classification commonly presents accuracy metrics measured on the same dataset-source incorporating the same diagnostic classes. Relying on those experimental results when implementing an automated classifier in clinical practice may be precarious, as an end-user may take images with a different camera, on patients with different skin types, with different class distributions - and even disease classes the network has not encountered before. For these reasons a classifier with a fixed set of diagnoses may fail in unexpected ways which would go unnoticed if the output is merely a probability of specific diagnoses.
Neural networks trained for classification by design are limited to predict classes they have seen during the training period. Currently, to our knowledge, no available dataset comes close to encompass all clinically possible classes. Further, class definitions of medical entities may change over time with new biologic insights. The CBIR method described herein shows that classifiers knowing only 3 classes are able to generalise better to a new dataset with 8 classes than their softmax based predictions (Table \ref{tbl:mapdiff}). The highest accuracy can still be obtained through finetuning a network on the target data source (blue lines in Figure \ref{fig:diffset}), but such a re-training period may not be feasible when retrieval data-sources are not accessible for training due to data protection regulations or lack of machine learning resources.

In contrast to decision support systems with a fixed performance and cutoff that needs to undergo clinical testing \cite{Monheit2011_melafind}, CBIR as a dynamic, and potentially vendor-independent, decision support system may be easier to expand and update in practice with growing search datasets and improved models.

\subsection{Limitations}
As the results from a previous study by Kawahara et al.\cite{Kawahara2018_7Point} were not public until the end of our experiments we did not perform a sample size calculation, so this work needs to be regarded an exploratory pilot study. We trained the ResNet-50 architecture on the datasets with reasonable effort on fine-tuning, data augmentation and hyperparameter tuning, but did not pursue maximum classification accuracy. Therefore, achievable values may be higher as shown by \cite{Han2018}, but we expect a better classifier using a larger image dataset to improve CBIR in a similar way.
All data herein is suffering from selection bias (they were found worthwhile to be photographed by a physician) and verification bias. A user-focused and prospective analysis of such a decision support will be able to give more insight in clinical applicability.

Document retrieval studies usually use a different set of metrics where mean average precision is defined differently. We chose the used metrics and definitions to reflect clinically meaningful outcomes rather than retrieval performance.

\section{Conclusion}
In this work we show that automated retrieval of few visually similar dermatoscopic images approximate accuracy of softmax-based prediction probabilities. Further, CBIR may improve performance of trained networks in new sets and unseen classes when there is no possibility of fine-tuning of a network on new data.


\newpage
\clearpage
\includepdf[pages=-]{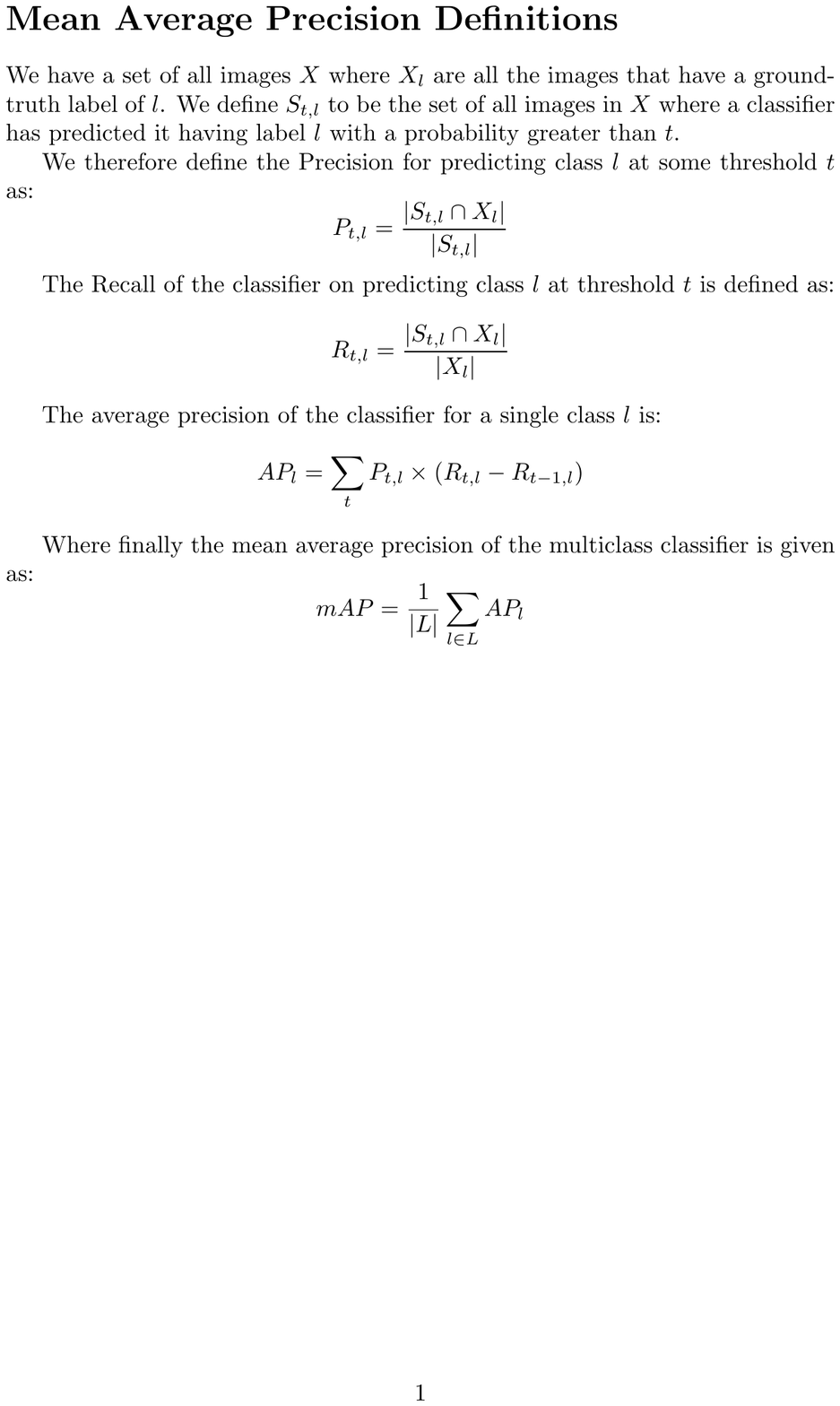}

\end{document}